\documentclass[twoside,leqno,twocolumn]{article}

\usepackage[letterpaper]{geometry}

\usepackage{ltexpprt}
\usepackage{hyperref}
\usepackage{microtype}
\usepackage{graphicx}
\usepackage{caption}
\usepackage{subcaption}
\usepackage{stackengine}
\newcommand\xrowht[2][0]{\addstackgap[.5\dimexpr#2\relax]{\vphantom{#1}}}
\usepackage{multirow}
\usepackage{arydshln}
\usepackage{amsmath}
\usepackage{amssymb}
\usepackage{mathtools}
\usepackage{xcolor}         % colors

\title{\Large Time-Transformer: Integrating Local and Global Features for Better \\ Time Series Generation}
\author{Yuansan Liu \thanks{University of Melbourne, Melbourne, VIC, Australia yuansanl@student.unimelb.edu.au} \and Sudanthi Wijewickrema \thanks{University of Melbourne, Melbourne, VIC, Australia \{sudanthi.wijewickrema, christofer.bester, sjoleary, baileyj\}@unimelb.edu.au} \and Ang Li \thanks{Academy of Military Sciences, Beijing, China angli.cs@outlook.com} \and Christofer Bester \footnotemark[2]\and Stephen O'Leary \footnotemark[2]\and James Bailey \footnotemark[2]}
\date{}
\begin{document}
\newcommand\relatedversion{}

\maketitle
\fancyfoot[R]{\scriptsize{Copyright \textcopyright\ 2024 by SIAM\\
Unauthorized reproduction of this article is prohibited}}

\begin{abstract} \small\baselineskip=9pt
  Generating time series data is a promising approach to address data deficiency problems. However, it is also challenging due to the complex temporal properties of time series data, including local correlations as well as global dependencies. Most existing generative models have failed to effectively learn both the local and global properties of time series data. To address this open problem, we propose a novel time series generative model named `Time-Transformer AAE', which consists of an adversarial autoencoder (AAE) and a newly designed architecture named `Time-Transformer' within the decoder. The Time-Transformer first simultaneously learns local and global features in a layer-wise parallel design, combining the abilities of Temporal Convolutional Networks and Transformer in extracting local features and global dependencies respectively. Second, a bidirectional cross attention is proposed to provide complementary guidance across the two branches and achieve proper fusion between local and global features. Experimental results demonstrate that our model can outperform existing state-of-the-art models in 5 out of 6 datasets, specifically on those with data containing both global and local properties. Furthermore, we highlight our model's ability to handle this kind of data via an artificial dataset. Finally, we show how our model performs when applied to a real-world problem: data augmentation to support learning with small datasets and imbalanced datasets.
\end{abstract}

\section{Introduction}
  Automatically synthesizing realistic data assists in solving real-world problems when access to real data is limited and manual generation is impractical. Deep generative models have shown great success in dealing with this problem of data deficiency in computer vision and natural language processing during the last decade. Numerous models have been introduced to address downstream tasks such as image in-painting \cite{pathak2016ganinpaint}, text to image translation \cite{zhang2016stackgan}, and automated captioning \cite{guo2017leakgan}.

  Although data deficiency is similarly crucial in the time series domain, there exist relatively few generative models to tackle this problem. This is due to the fact that the generated data is required to share a similar global distribution with the original time series data and also preserve its unique temporal properties.
  % As such, generative models for time series data, especially those universally applicable to different types of time series data are relatively rare.
  Some existing works utilize Generative Adversarial Networks (GANs) \cite{goodfellow2014gan} for time series generation and most of them address the temporal challenges using Recurrent Neural Networks (RNNs) such as Long Short Term Memory (LSTM) \cite{esteban2017rcgan,yoon2019timegan,pei2021rtsgan}. There are also approaches that use Variational Autoencoder (VAE) \cite{kingma2014vae} as the basic framework to generate time series data \cite{fabius2014vrae,desai2021timevae}. However, none of these works have fully succeeded in efficiently learning both local correlation and global interaction, which is critical for time series processing.

  Recently, Transformer based models have been successful in learning global features for different types of data including time series \cite{raffel2019transtext,dosov2020vit,zerveas2021transtsrl,chen2021mobileformer,chen2022s2tnet}. On the other hand, models based on Convolutional Neural Networks (CNNs) have shown their strength in extracting local patterns with their filters \cite{howard2017mobile,yamashita2018cnnrwv,liu2019fgcnn}. Temporal Convolutional Networks (TCNs), consisting of dilated convolutional layers \cite{oord2016wavenet}, preserve the original local processing capability of CNNs, but also have an enhanced ability to learn temporal dependencies in sequential data \cite{bai2018tcn}. This makes them appropriate for time series modeling. Therefore, it is natural to combine Transformer and TCN together to learn better time series features. Some previous works use sequential combinations of these for time series tasks \cite{lin2019atttcnmedts,cao2021tcnrsa}. However, such sequential designs do not consider the interaction between local and global features inherent in these datasets. Thus, it is critical to learn different levels of features separately to preserve their independence, and then, integrate them with an appropriate method.

  Motivated by the above observations and analysis, we propose a novel time series generative model named `Time-Transformer AAE'. Specifically, we first use the Adversarial Autoencoder (AAE) \cite{makhzani2015aae} as the generative framework. The Time-Transformer is designed as part of the decoder to efficiently and  effectively learn and integrate both local and global features. In each Time-Transformer block, the temporal properties are learnt by both a TCN layer and a Transformer block. They are then connected through a bidirectional cross attention block to fuse local and global features. This layer-wise parallelization, along with bidirectional interaction, combines the advantages of the TCN and Transformer models: the efficiency of the TCN to extract local features, as well as the Transformer’s ability in building global dependency. We evaluate our proposed Time-Transformer AAE on different types of time series data from both artificial and real-world datasets. Experiments show that the proposed model surpasses the existing state-of-the-art (SOTA) models in addressing the time series generation task. Furthermore, we also show our model's effectiveness on downstream tasks including small dataset augmentation and imbalanced classification. To summarize, our contributions are as follows:
  \vspace{-3pt}
  \begin{itemize}
      % \item We propose a new time series generative model called Time-Transformer AAE, which effectively combines the advantages of TCN and Transformer in extracting local and global patterns respectively.
      \item We introduce the Time-Transformer module that simultaneously learns local and global features in a layer-wise parallel design, and facilitates interaction between these two types of features by performing feature fusion in a bidirectional manner.
      \item By inserting the proposed module into the decoder of the Adversarial Autoencoder, we present a novel time series generative model, Time-Transformer AAE, which efficiently and effectively generates high quality synthetic time series data containing both global and local features.
      \item We show empirically that the proposed Time-Transformer AAE can generate better synthetic time series data with respect to different benchmarks, when compared to SOTA methods.
      \item Through comprehensive experiments, we demonstrate the efficiency and effectiveness of our model in handling time series data containing both global and local features, and subsequently, addressing data deficiency issues.
  \end{itemize}

 \section{Related Works} \label{literw}
  \subsection{Time Series Generation:}
   Deep generative models (DGMs) have gained increasing attention since their introduction. The variational autoencoder (VAE) uses the Bayesian method to learn latent representations and turn the classic autoencoder into a generative model \cite{kingma2014vae}. Generative adversarial networks (GANs) introduce an adversarial approach to shape the output distribution in order to generate realistic fake data \cite{goodfellow2014gan}. The adversarial autoencoder (AAE) combines the previous two models together \cite{makhzani2015aae}. It uses the adversarial training procedure to perform variational inference in the VAE. Numerous models have been designed based on these basic generative frameworks that have shown superior performance in image and text processing \cite{oord2016pixrnn,pathak2016ganinpaint,zhang2016stackgan,karras2017progan,arjovsky2017wgan,he2018lagging,ahamad2019stgangp}.

   % Here, we review previous works in standard time series generation, where complete time series data is required. There also exist works investigating generative models for incomplete time series, which are popular variants of the standard ones. But these methods are outside the scope of this paper.

   Successes in processing graphs and text have led to the application of DGMs' in time series problems. Most of them use the GAN framework with modifications to incorporate temporal properties. The first of these, called C-RNN-GAN \cite{mogren2016crnngan}, directly uses the GAN structure with LSTM to generate music data. Recurrent Conditional GAN (RCGAN) \cite{esteban2017rcgan} uses a basic RNN as its backbone, and auxiliary label information as the condition to generate medical time series. Since its introduction, a number of works have utilized similar designs to generate time series data in various fields including finance, medicine and the internet \cite{hartmann2018eeggan,Koochali2019forgan,wiese2020quantgan,smith2020tsgan,lin2020gannetwork}. TimeGAN \cite{yoon2019timegan} introduces additional embedding function and supervised loss to generate universal time series. Real-world Time Series GAN (RTSGAN) \cite{pei2021rtsgan} combines WGAN \cite{arjovsky2017wgan} and autoencoder, and achieves good performance on real-world data generation. PSA-GAN \cite{jeha2022psagan} utilizes the progressively growing architecture \cite{karras2017progan} to generate long time series. To generate both regular and irregular sampled time series, GT-GAN \cite{jeon2022gtgan} combines Neural Differential Equations and DGMs together for the first time. COSCI-GAN \cite{seyfi2022coscigan} focuses on multivariate time series, and uses a central discriminator to learn better inter-channel dynamics. On the other hand, VAE has also been used for this task. \cite{fabius2014vrae} designed a recurrent VAE to synthesize time series data. TimeVAE \cite{desai2021timevae} implements an intepretable temporal structure together with VAE, and achieves SOTA results on universal time series generation. CR-VAE \cite{li2023causal} incorporates the causal mechanism into time series generation by learning a Granger causal graph. Recently, the diffusion model has also been utilized together with TCNs to generate audio data \cite{kong2020diffwave}. In comparison to traditional DGMs, TimeGCI uses the contrastive learning framework to generate time series and achieves good results \cite{jarrett2021timeci}. These works learn the overall distribution of the training set and generate synthetic time series within the same distribution.

   On the other hand, another line of works focus on step-wise generation: They build generative models to infer each time step on the previous ones. Some early works modify the basic RNN using variational inference and achieve good results with regard to log-likelihood \cite{chung2015vrnn,fraccaro2016sequential}. Some later works combine this idea with TCN \cite{lai2018swavenet,aksan2019stcn}. In recent years, diffusion models are used to achieve SOTA results in this area \cite{rasul2021arddpm,li2022generative}. Though different to our task, these works provide an instructive direction for applying generative models on the time series forecasting task.

  \subsection{Temporal Convolutional Networks and Transformer:}
   Temporal Convolutional Networks (TCNs) \cite{bai2018tcn} use dilated causal convolutions in WaveNet \cite{oord2016wavenet} to encode temporal patterns and avoid information leakage from future to past. TCN models have been used for sequential data such as time series and are able to successfully extract local features \cite{sen2019deepglo,zhao2019tcntraffic,hewage2020tcnweats,deldari2021tscpdcpc}. Transformer \cite{vaswani2017tranformer} implements a self-attention mechanism on the entire data to learn global interactions between each point, which enhances ability to learn long range dependence and improves its performance on machine translation tasks. Its variants also achieve success on language, vision and time series tasks \cite{raffel2019transtext,dosov2020vit,zerveas2021transtsrl,chen2022s2tnet}. Some works combine these two types of models sequentially (place a Transformer either before or after a TCN) to take advantage of both their capabilities \cite{lin2019atttcnmedts,cao2021tcnrsa}. However, this design assumes dependencies exist between the models. However, a parallel structure like our model is necessary if different levels of features need to be learnt separately to preserve their independence.

 % \section{Methods}
  % In this section, we first describe the problem. Then, we introduce the proposed Time-Transformer AAE architecture. Afterwards, we discuss details of the model using one Time-Transformer block.

\section{Problem Formulation}
   Generally, multivariate time series generation via GAN-like models involves training a model to approximate a function $\mathcal{F}: \mathcal{P} \rightarrow \mathcal{P}_d$ which maps an arbitrary prior distribution $\mathcal{P}$ to the real data distribution $\mathcal{P}_d$, so that the model can produce realistic synthetic data $X' \sim \mathcal{P}_d$ based on any samples $s$ drawn from $\mathcal{P}$. For time series data $X \in \mathbb{R}^{T \times c}$, where $T$ is maximum observable time and $c$ is number of channels, we define the local patterns and the global dependencies as:
   \begin{align}
       \mathcal{L} &= p(X_{k:t-1}) \\
       \mathcal{G} &= p(X_t\;|\; X_{1:t-1})
   \end{align}
   where $1<k<t$ and $p$ is the density. They are derived from previous researches \cite{yoon2019timegan,yang2022local} with adaptations to align with our requirements. Thus, $X$ contains both $\mathcal{L} and \mathcal{G}$ and function $\mathcal{F}$ must learn them simultaneously:
   \begin{equation} \label{func}
       \mathcal{F}(p) = \mathcal{I}(\mathcal{L}, \mathcal{G}), 1<k<t
   \end{equation}
   where $\mathcal{I}$ is an integration function that combines local and global features together.

   The goal of our work is to design a model to approximate $\mathcal{F}$. It should extract both local and global features and properly combine them to generate realistic and useful synthetic time series data that can be used for downstream real-world tasks.

 \section{Time-Transformer AAE}
   \subsection{Overview:}
\begin{figure}[ht]
        \centering
        \scalebox{0.5}{%
        \begin{tabular}{c}
            \begin{subfigure}[b]{0.9\textwidth}
                \centering
                \includegraphics[scale=0.72]{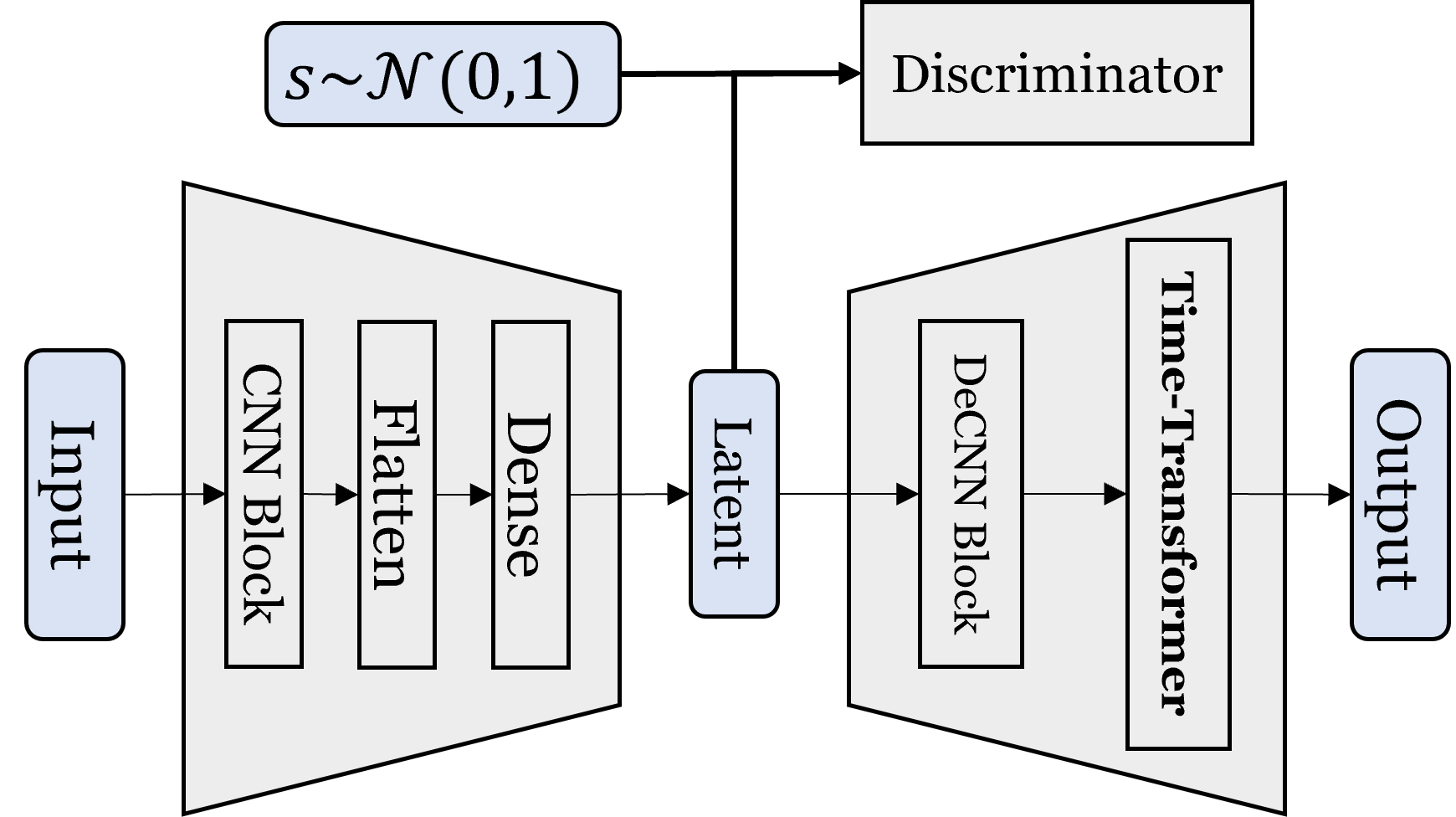}
                \caption{Overview of Time-Transformer AAE}
                \label{overvw}
            \end{subfigure} \\
            \hdashline \\
            \begin{subfigure}[b]{0.9\textwidth}
                \centering
                \includegraphics[scale=0.65]{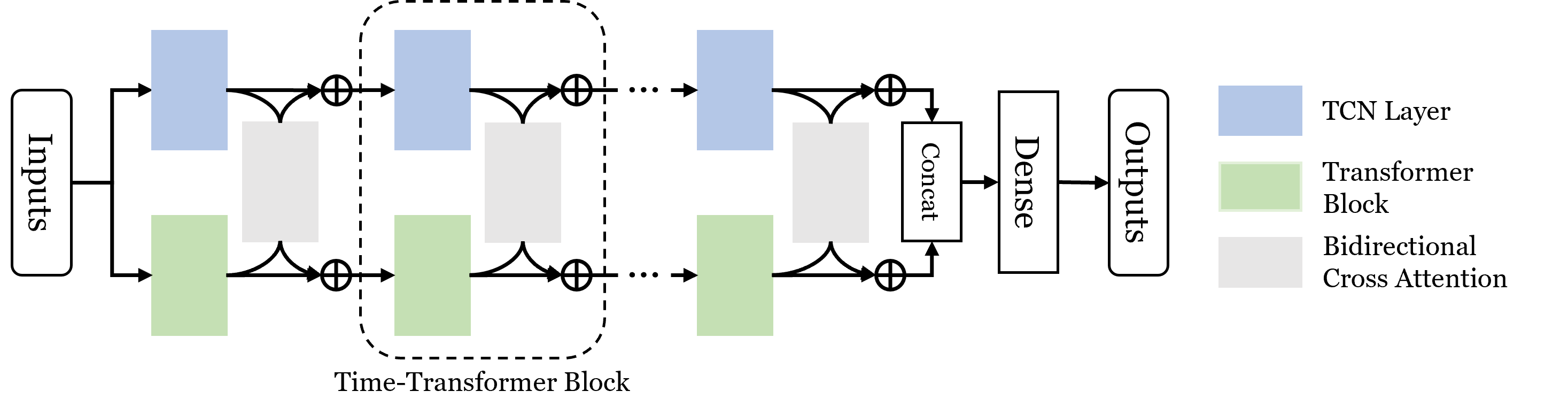}
                \caption{Time-Transformer Structure}
                \label{ttrans}
            \end{subfigure}
        \end{tabular}}
        \caption{Time-Transformer AAE}
    \end{figure}

    We use Adversarial Autoencoder (AAE) as our generation framework due to the potential of extending it to supervised and semi-supervised learning settings \cite{makhzani2015aae}. To apply AAE for time series generation, we first use Convolutional Neural Networks (CNNs) as the basis of the AAE, and then insert the Time-Transformer (see Figure \ref{ttrans}) into the decoder of the AAE. The overall structure is shown in Figure \ref{overvw}.

    Here, we use Gaussian as the prior distribution $\mathcal{P}=\mathcal{N}(0,1)$ and insert the Time-Transformer after a De-Convolutional block to approximate the $\mathcal{F}$ in \eqref{func} (This design is to achieve an optimal equilibrium between efficiency and effectiveness. We provide details in \cite{liu2023timetransformer}). The De-Convolutional block will first reconstruct a prototype of the time series, and then the Time-Transformer aims at learning the detailed local/global features through feedback from reconstruction loss during training. Finally, the model will be able to synthesize realistic data in the generation step.

    Within the Time-Transformer, we use a layer-wise parallelization to combine the Temporal Convolutional Networks (TCNs) and the Transformer. The learnt prototype from the De-Convolutional block is passed into a TCN layer and a Transformer block simultaneously. The TCN learns the local features of the time series, and the Transformer finds the global patterns of the data. The learnt results then go through a cross-attention block to fuse with each other bidirectionally. At the end of this parallel structure we concatenate the outputs from both sides, and use a fully-connected layer to map them into the expected dimension ($t \times c$) and reshape into the original time series dimensions.

   \subsection{Time-Transformer:}
\begin{figure}[ht]
        \centering
        \includegraphics[scale=0.4]{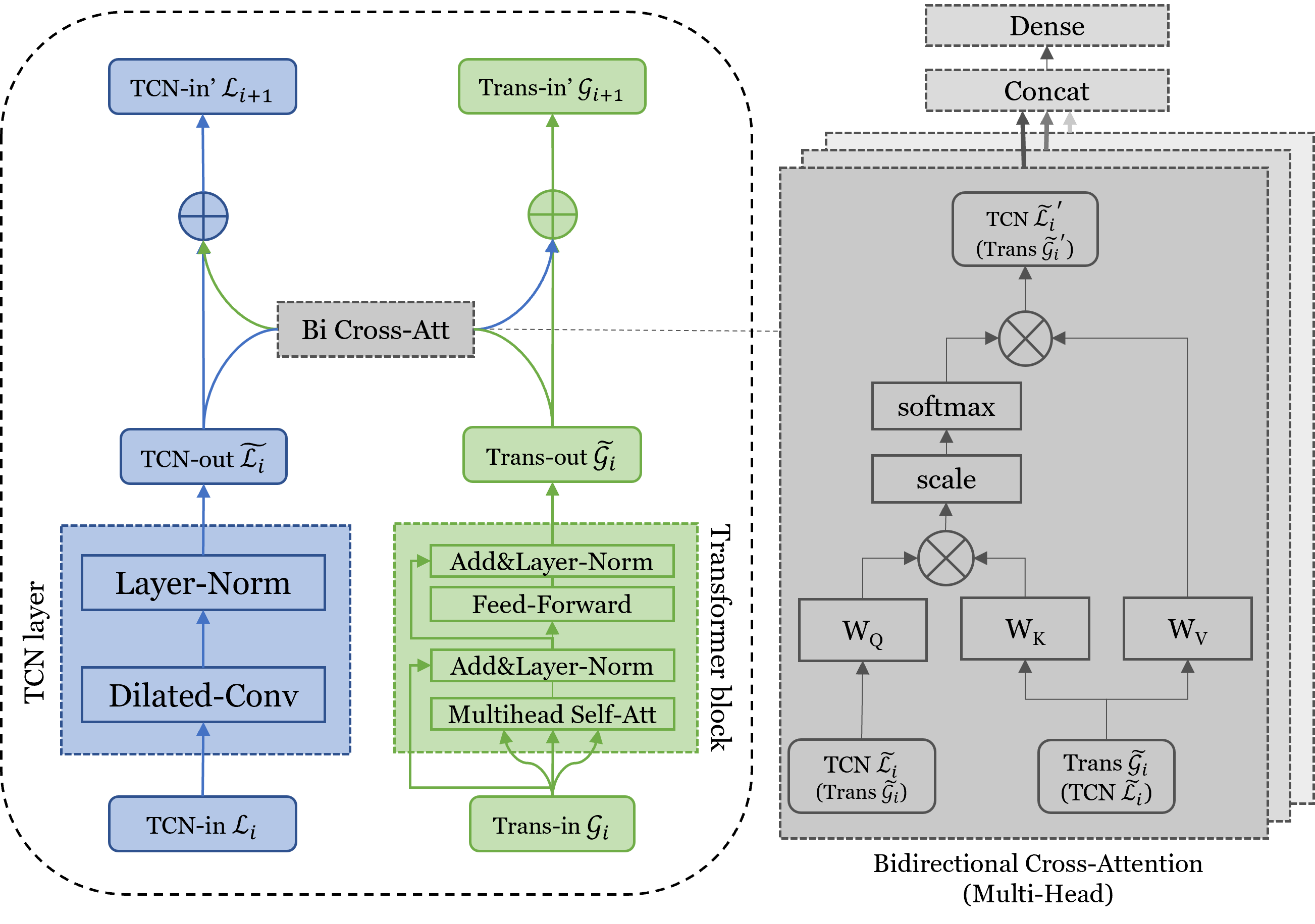}
        \caption{Detailed design of the Time-Transformer block}
        \label{tfblock}
    \end{figure}

    The Time-Transformer consists of several Time-Transformer blocks, which have two key differences to the standard TCN layers and Transformer blocks: 1) a layer-wise parallel design to combine local-window self-attention and depth-wise convolution and 2) bidirectional cross attention over the two branches. Figure \ref{tfblock} shows the details of the Time-Transformer block.

    \textbf{Layer-wise Parallelization}: The Mobile-Former \cite{chen2021mobileformer} combines MobileNet (a light-weight CNN) and Transformer in parallel, and achieve better performance than sequentially combined models on image classification and object detection tasks. Inspired by this, we also combine TCN and Transformer in a parallel manner to learn $\mathcal{L}$ and $\mathcal{G}$ in \eqref{func} simultaneously. However, instead of simply combining the entire TCN and Transformer, we only use one layer from the TCN and one block from the Transformer in each Time-Transformer block.

    As shown in Figure \ref{tfblock}, each Time-Transformer block has TCN and Transformer (Trans) streams parallel to each other. Inside the TCN stream we have a dilated convolutional layer with post layer normalization, which make up one hidden layer of the TCN model. The inputs and outputs of each block (TCN-in and TCN-in' respectively in Figure \ref{tfblock}) are local features including those learnt directly from the last/current TCN layer (TCN-out in Figure \ref{tfblock}) and the fusion of the local and global features. The Transformer block has a Multi-head Self Attention layer and a Feed-forward Network, both with post layer normalization. Similarly, the inputs/outputs (Trans-in and Trans-in' respectively in Figure \ref{tfblock}) of this side are global features from last/current Transformer block together with a feature fusion.

   \textbf{Bidirectional Cross Attention}:
   The bidirectional cross attention block contains two cross attention mechanisms, aims to build interactions between two parallel branches, and thus fuses local and global features to function as the $\mathcal{I}$ in \eqref{func}. As illustrated on the right hand side of Figure \ref{tfblock}, the output of the TCN layer $\widetilde{\mathcal{L}_i}\in \mathbb{R}^{t \times c}$ and the output of the Transformer block $\widetilde{\mathcal{G}_i}\in \mathbb{R}^{t \times c}$ interact in a mutual manner within the cross attention block, where it bidirectionally fuses the local feature $\widetilde{\mathcal{L}_i}$ and the global feature $\widetilde{\mathcal{G}_i}$. Specifically, the TCN features are updated through a residual connection with the attention matrix to obtain $\mathcal{L}_{i+1}$:
    \begin{equation} \label{xi1}
      \mathcal{L}_{i+1} = \widetilde{\mathcal{L}_i} +  \mathcal{A}_{\widetilde{\mathcal{G}_i} \rightarrow \widetilde{\mathcal{L}_i}}\cdot \mathcal{G}_iW_{ev}
   \end{equation}

   where $W_{ev}$ is the learnable parameter for the value embedding layer, and $ \mathcal{A}_{\widetilde{\mathcal{G}_i} \rightarrow \widetilde{\mathcal{L}_i}}$ is the affinity matrix from Trans to TCN which can be calculated with matrix multiplication and a softmax function:
     \begin{equation} \label{Axz}
      \mathcal{A}_{\widetilde{\mathcal{G}_i} \rightarrow \widetilde{\mathcal{L}_i}} = softmax(\frac{\widetilde{\mathcal{L}_i}W_{eq}\cdot (\widetilde{\mathcal{G}_i}W_{ek})^{T}}{\sqrt{c}})
   \end{equation}

   where $W_{eq}$ and $ W_{ek}$ are learnable parameters of two linear projection layers. Similarly, the updated Trans feature map $\mathcal{G}_{i+1}$ is defined as:
     \begin{align} \label{zi1}
      \mathcal{G}_{i+1} = &\widetilde{\mathcal{G}_i} +  \mathcal{A}_{\widetilde{\mathcal{L}_i} \rightarrow \widetilde{\mathcal{G}_i}}\cdot \mathcal{L}_iW_{dv} \\
      \mathcal{A}_{\widetilde{\mathcal{L}_i} \rightarrow \widetilde{\mathcal{G}_i}} = & softmax(\frac{\widetilde{\mathcal{G}_i}W_{dq}\cdot (\widetilde{\mathcal{L}_i}W_{dk})^{T}}{\sqrt{c}})
   \end{align}

     where $W_{dq}$, $ W_{dk}$, and $W_{dv}$ are learnable parameters of three linear projection layers. $ \mathcal{A}_{\widetilde{\mathcal{L}_i} \rightarrow \widetilde{\mathcal{G}_i}}$ is the affinity matrix from TCN to Trans

 \section{Experiments}
  \subsection{Datasets:}
   We use six datasets to evaluate our model. The first three datasets have been used in several previous works \cite{yoon2019timegan,pei2021rtsgan,desai2021timevae}. Hence, they are referred to as \textit{preliminary datasets}. The next three datasets contain time series with both local and global patterns. They are used to evaluate the model's ability to learn both types of features, and are termed \textit{local-global datasets}:

    \textbf{Sine\_Sim}: 5000 samples of 5-dimensional sinusoidal sequences: $x_t^{(d)}=\alpha sin(2\pi f t + \varphi)$ where $d \in [1,5], \alpha \sim \mathcal{U}[1,3], f\sim \mathcal{U}[0.1, 0.15]$ and $\varphi \sim \mathcal{U}[0, 2\pi]$

    \textbf{Stock}: 3686 samples of stock-price from historical Google daily stocks. Each sample has 6 features correlated with each other.

    \textbf{Energy}: The UCI Appliances energy prediction dataset contains 19735 time series data with high dimensionality (28) and correlated features.

   \textbf{Sine\_Cpx}: Another 5000 samples of more complex 5-dimensional sinusoidal sequences. They are simulated using the sum of standard sinusoidal waves $x_t^{(d)}=\alpha_1 sin(2\pi f_1 t + \varphi_1) + \alpha_2 sin(2\pi f_2 t + \varphi_2) + \alpha_3 sin(2\pi f_3 t + \varphi_3)$ where $\alpha_i \sim \mathcal{U}[1,3], f_i\sim \mathcal{U}[0.1, 0.15], \varphi_i \sim \mathcal{U}[0, 2\pi], i=1,2,3$ and $\alpha_i \neq \alpha_j, f_i \neq f_j, \varphi_i \neq \varphi_j$, when $i \neq j$. Thus, the sequences demonstrate global sine waves and local patterns in each dimension.

   \textbf{Music}: A 2-dimensional time series waveform audio file of classical music 'Canon in D' converted from MP3, sampled at 8000Hz frequency with double channels, resulting in 2845466 time-steps. The music data exhibits local patterns and global seasonality. A 10000-step segment (2'05" to 2'06") is selected for our experiments.

   \textbf{ECochG}: A medical dataset of historical patient data from cochlear implant surgeries \cite{wijewickrema2022automatic}. Each instance contains a univariate time series that represents a patient's inner ear response during the surgery. In general, each time series is a stochastic trend globally, with some local drops. More details of this dataset can be found in \cite{liu2023timetransformer}.

  \subsection{Baseline Models \& Benchmarks:}
   We select previous models, TimeVAE, RTSGAN, and TimeGAN \cite{desai2021timevae,pei2021rtsgan,yoon2019timegan}, as baselines based on their relevance, accessibility of the code, executability of the code, and performance of the model\footnote{Models aligned with our task but failed any of these criteria are excluded. E.g., PSA-GAN's released codes are not executable}. We use their original settings as they have already been optimised by their authors. Details of our model\footnote{The code is available at \url{https://github.com/Lysarthas/Time-Transformer}} (hyper-parameters and data pre-processing) can be found in \cite{liu2023timetransformer}. To evaluate these generative models, we use a range of metrics and consider several desiderata for the generated time series: They should (1) be distributed close to the original data and different from each other, (2) be indistinguishable from the original data, and (3) preserve the original temporal properties in order to be used for downstream tasks.

   \textbf{Visualisation} is an indicator of (1) showing the distributions over the original and generated data. We flatten the temporal dimension and use t-SNE \cite{laurens2008tsne} plots for 2-dimensional space for visualisation.

   \textbf{Fréchet Inception Distance} (FID) evaluates quality of generated data for GAN type models. Following\cite{hartmann2018eeggan,smith2020tsgan,jeha2022psagan}, we replace the Inception model of the original FID with a recent SOTA time series representation learning method TS2Vec \cite{yue2021ts2vec}. FID score provides a quantitative assessment of (1).

   \textbf{Discriminative Score} measures indistinguishability of the generated data. We follow protocol from \cite{yoon2019timegan} when defining this and next benchmarks. We first train a post-hoc sequential classifier (a 2-layer LSTM) to distinguish original and generated data. Then we get test accuracy $acc_{te}$ from a held-out test set. Good generated time series data should be indistinguishable from real data (accuracy close to 0.5). Thus, we use $|acc_{te}-0.5|$ as the Discriminative Score to measure (2).

   \textbf{Predictive Score} reflects how well the generated data inherits the predictive properties of the original data. We train a post-hoc sequential predictor (a 2-layer LSTM) to do one-more-step prediction, and calculate the mean absolute error (MAE). Here, we use the `Train on Synthetic, Test on Real' (TSTR) technique introduced by \cite{esteban2017rcgan}. The predictor should be able to learn from the generation and predict the real data. The Predictive Score provides a quantitative assessment of (3).

  \subsection{Experimental Results} \label{res}
   % In this section, we first report the performance of the models with respect to the aforementioned benchmarks. For the three scores, we execute the experiment several times to get the averages and confidence intervals (95\% confidence). Afterwards, we conduct some further studies using several real-world datasets.

   \subsubsection{Visualisation \& FID:}
\begin{figure}[ht]
     \centering
     \includegraphics[scale=0.3]{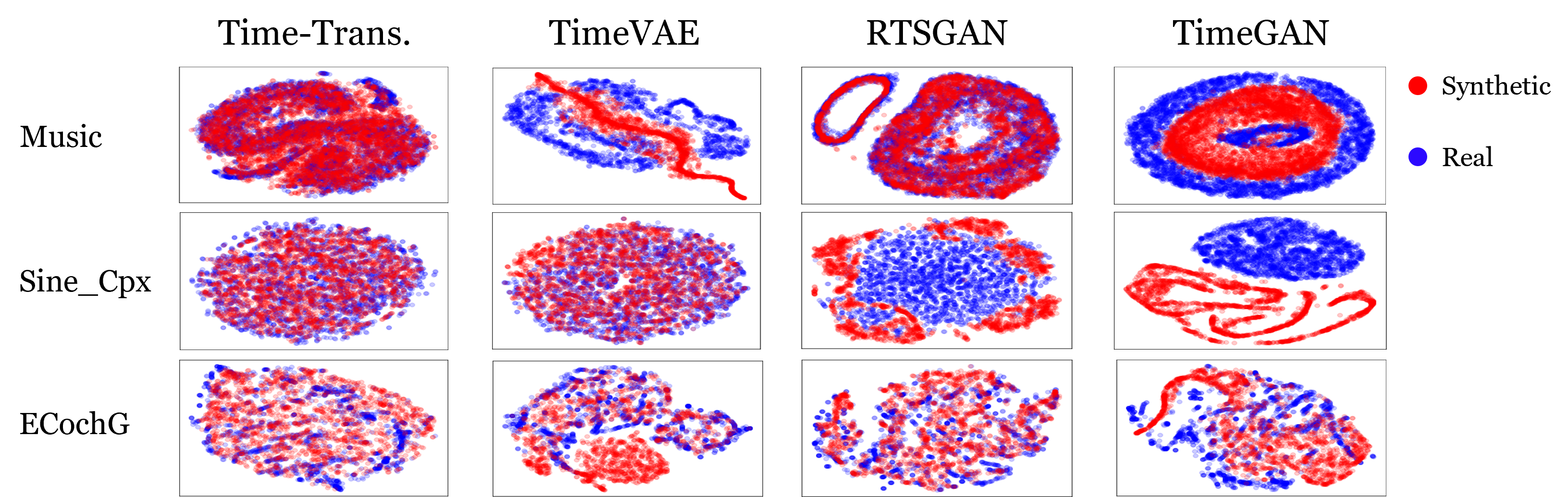}
     \caption{t-SNE visualisations of models (columns) on 3 datasets (rows). The blue dots represent original data and the red dots represent synthetic data.}
     \label{tsne}
    \end{figure}

    Figure \ref{tsne} shows visualization results of three local-global datasets (t-SNE plots for preliminary datasets can be found in \cite{liu2023timetransformer}). Each row contains plots of one dataset and each column represents a model. The blue and red dots represent original and generated data points respectively. It is obvious that our proposed Time-Transformer AAE (denoted as \textbf{Time-Trans} hereafter) consistently produces synthetic data closely distributed to the original data. TimeVAE and RTSGAN also have good distributions for most datasets. Table \ref{fid} shows the FID scores of each model on each dataset. Our proposed Time-Trans obtains the best scores for all the datasets except ‘Stock’ where RTSGAN is slightly better. For those datasets where the models show similar visualisation results (e.g. Time-Trans and TimeVAE on `Sine\_Cpx'), we can use the FID scores to quantitatively evaluate them.

    \begin{table}[ht]
     \centering
     \caption{FID score: lower scores are better}
     \scalebox{0.7}{%
     \begin{tabular}{ccccc}
         \hline\xrowht{5pt}
         Models & \textbf{Time-Trans} & TimeVAE & RTSGAN & TimeGAN\\ \hline\hline\xrowht{5pt}
         Sine\_Sim & \textbf{0.283±0.023} & 0.766±0.027 & 3.175±0.142 & 4.572±0.153 \\
         Stock & 0.273±0.035 & 0.303±0.037 & \textbf{0.254±0.031} & 0.646±0.072 \\
         Energy & \textbf{0.983±0.039} & 1.598±0.067 & 6.292±0.265 & 2.506±0.126 \\
         Music & \textbf{0.395±0.013} & 13.415±0.937 & 0.957±0.095 & 1.894±0.121 \\
         Sine\_Cpx & \textbf{1.502±0.062} & 3.486±0.170 & 7.696±0.251 & 14.255±0.491 \\
         ECochG & \textbf{0.348±0.024} & 5.197±0.237 & 0.527±0.062 & 0.674±0.098 \\ \hline
     \end{tabular}}
     \label{fid}
    \end{table}

   \subsubsection{Discriminative Score \& Predictive Score:}
    As per the Discriminative Scores shown in Table \ref{discr},\begin{table*}[ht]
     \centering
     \caption{Discriminative score: $|acc_{te}-0.5|$, lower scores are better}
     \scalebox{0.87}{%
     \begin{tabular}{ccccc}
         \hline \xrowht{5pt}
         Models & \textbf{Time-Trans} & TimeVAE & RTSGAN & TimeGAN\\ \hline\hline\xrowht{5pt}
         Sine\_Sim & \textbf{0.131±0.021} & 0.217±0.015 & 0.489±0.007 & 0.485±0.008 \\
         Stock & 0.463±0.023 & 0.476±0.050 & \textbf{0.374±0.022} & 0.481±0.033 \\
         Energy & \textbf{0.496±0.005} & 0.499±0.002 & 0.499±0.001 & 0.499±0.001 \\
         Music & \textbf{0.160±0.054} & 0.495±0.008 & 0.489±0.009 & 0.494±0.006 \\
         Sine\_Cpx & \textbf{0.168±0.041} & 0.471±0.016 & 0.489±0.031 & 0.500±0.000 \\
         ECochG & \textbf{0.103±0.012} & 0.474±0.016 & 0.405±0.004 & 0.424±0.015 \\ \hline
      \end{tabular}}
      \label{discr}
    \end{table*}

    \begin{table*}[h!]
     \centering
     \caption{Predictive score: 10$\times$MAE, lower scores are better}
     \scalebox{0.87}{%
     \begin{tabular}{ccccccc}
        \hline \xrowht{5pt}
        Model & Oracle & \textbf{Time-Trans} & TimeVAE & RTSGAN & TimeGAN\\ \hline\hline\xrowht{5pt}
        Sine\_Sim & 0.038±0.012 & \textbf{0.051±0.015} & 0.108±0.031 & 0.789±0.069 & 0.465±0.055 \\
        Stock & 0.015±0.003 & 0.048±0.004 & 0.080±0.003 & \textbf{0.036±0.003} & 0.095±0.007 \\
        Energy & 0.044±0.004 & 0.077±0.007 & \textbf{0.072±0.006} & 0.228±0.022 & 0.115±0.013 \\
        Music & 0.014±0.008 & \textbf{0.017±0.009} & 0.077±0.051 & 0.021±0.009 & 0.027±0.007 \\
        Sine\_Cpx & 0.029±0.008 & \textbf{0.032±0.006} & 0.055±0.015 & 0.070±0.006 & 0.527±0.025 \\
        ECochG & 0.011±0.009 & \textbf{0.013±0.006} & 0.087±0.008 & 0.176±0.021 & 0.042±0.006 \\ \hline
     \end{tabular}}
     \label{pred}
    \end{table*}
    our Time-Trans performs well for most datasets, and particularly excels in performance for the local-global datasets where the generations from our model show better indistinguishability than the other methods.

    We report ten times MAE as Predictive Score to avoid long decimal numbers while highlighting significant differences. An additional column `Oracle' shows results from the `Train on Real, Test on Real', which represent the theoretically best performances. Table \ref{pred} shows a similar trend:
    % \begin{table*}[h!]
    %  \centering
    %  \caption{Predictive score: 10$\times$MAE, lower scores are better}
    %  \scalebox{0.9}{%
    %  \begin{tabular}{ccccccc}
    %     \hline \xrowht{5pt}
    %     Model & Oracle & \textbf{Time-Trans} & TimeVAE & RTSGAN & TimeGAN\\ \hline\hline\xrowht{5pt}
    %     Sine\_Sim & 0.038±0.012 & \textbf{0.051±0.015} & 0.108±0.031 & 0.789±0.069 & 0.465±0.055 \\
    %     Stock & 0.015±0.003 & 0.048±0.004 & 0.080±0.003 & \textbf{0.036±0.003} & 0.095±0.007 \\
    %     Energy & 0.044±0.004 & 0.077±0.007 & \textbf{0.072±0.006} & 0.228±0.022 & 0.115±0.013 \\
    %     Music & 0.014±0.008 & \textbf{0.017±0.009} & 0.077±0.051 & 0.021±0.009 & 0.027±0.007 \\
    %     Sine\_Cpx & 0.029±0.008 & \textbf{0.032±0.006} & 0.055±0.015 & 0.070±0.006 & 0.527±0.025 \\
    %     ECochG & 0.011±0.009 & \textbf{0.013±0.006} & 0.087±0.008 & 0.176±0.021 & 0.042±0.006 \\ \hline
    %  \end{tabular}}
    %  \label{pred}
    % \end{table*}
    our model performs well in most datasets, particularly, in local-global datasets where our model's scores are close to Oracle. This indicates that our model has learnt most of the predictive properties of these datasets. Note that scores in our work differ to those published in related works because we calculate them in a slightly different way. We provide a detailed justification in \cite{liu2023timetransformer}. Apart from these six datasets, we also evaluated our model on longer time series data. The results show that our model consistently out-performs baselines with these longer time series (details can be found in \cite{liu2023timetransformer}).

   \subsubsection{Local \& Global Features Learning:}
\begin{figure}[ht]
        \centering
        \includegraphics[scale=0.41]{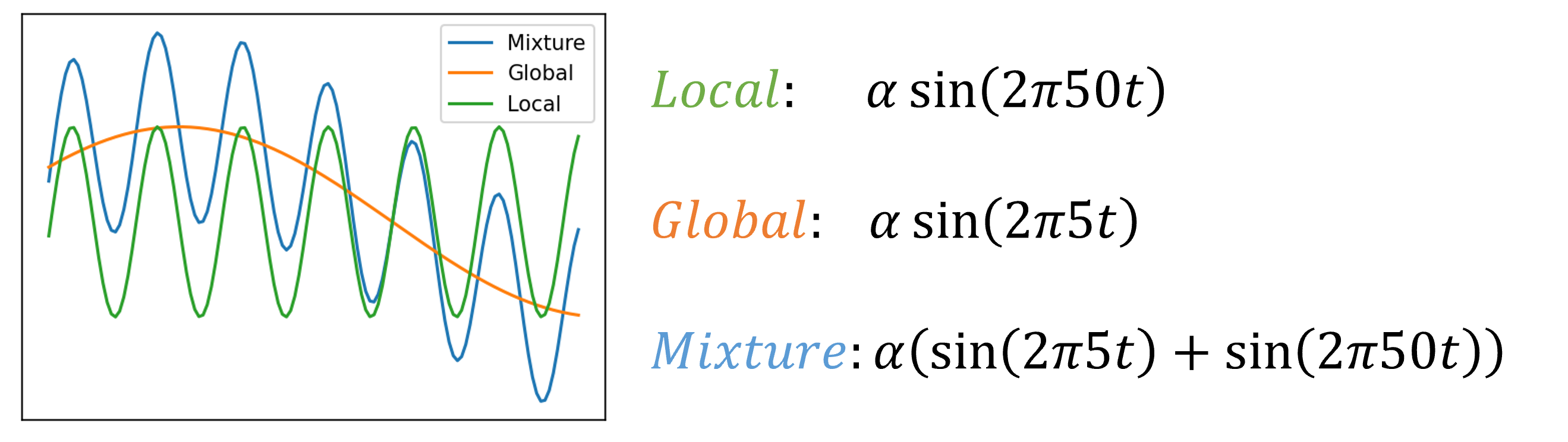}
        \caption{Mixture, Global and Local time series}
        \label{lgsample}
    \end{figure}

    We artificially simulate datasets to study the performance of the models with time series containing varying portions of global and local features. Figure \ref{lgsample} shows the underlying idea.
    The green and orange series represent local and global features respectively. They are generated with high (50Hz) and low (5Hz) frequencies with respect to the Fast Fourier Transform (FFT). By adding them together we get the blue series which, contains both global and local features (named `Mixture'). The formulae to generate local, global and mixture time series are alongside, where $\alpha \sim \mathcal{U}[1,3]$.

    For each of the three cases, we generate 5000 time series, each consisting of 128-step time steps (we use 128 to ensure global features are non-trivial). We thus obtain three different datasets, each containing 5000 time series.

\begin{figure}[ht]
        \centering
        \includegraphics[scale=0.42]{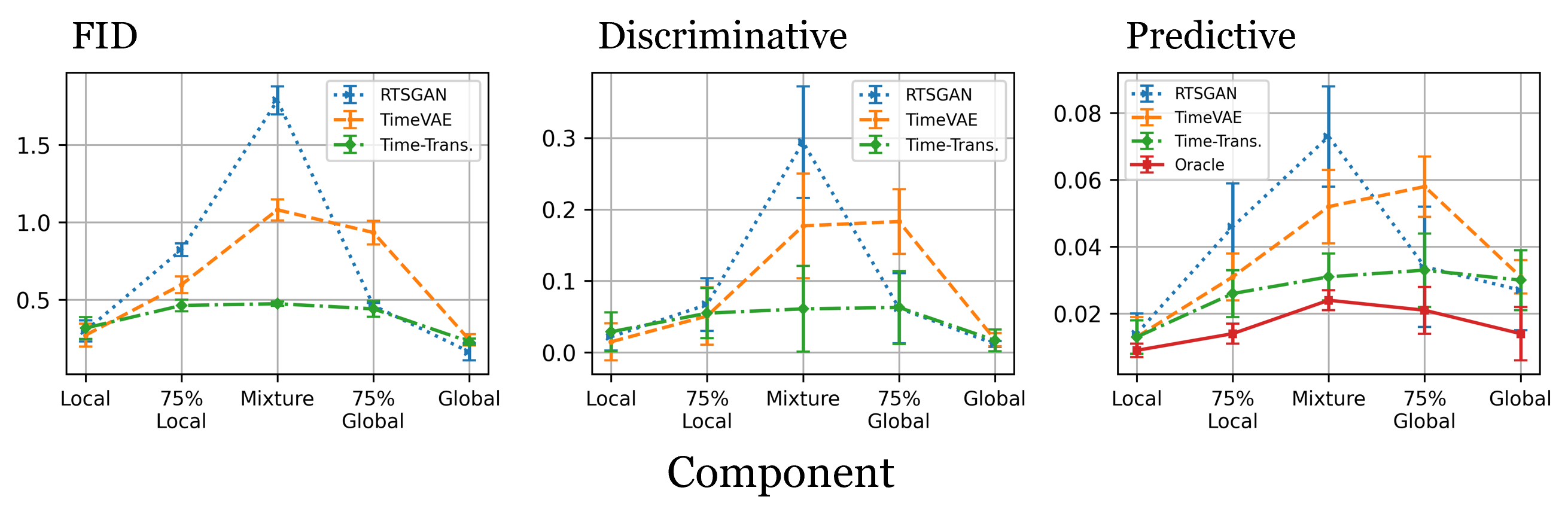}
        \caption{Results of Time-Trans(green), RTSGAN(blue), and TimeVAE(orange). Lower are better.}
        \label{locglb}
    \end{figure}

    We vary the portion of each part in Mixture to get two other series that contain 75\% local and 75\% global respectively. We train the generative models on these datasets and compare performance. Figure \ref{locglb} shows the above scores (mean score with standard deviation) of the three best models (RTSGAN, TimeVAE, and Time-Trans).
    When the data only contains one feature type, all models can generate comparatively good time series, but their performances decrease with mixed features. Among these models, ours shows the most consistent performance in all cases and has the best performance in mixture datasets with respect to all metrics, indicating its effectiveness in learning both local and global features. We provide detailed results in \cite{liu2023timetransformer}.

   \subsubsection{Model Application - Data Augmentation:} \label{apply}
    We evaluate our model on imbalanced classification using the ECochG dataset and three real-world datasets from the UCR archive \cite{UCRArchive2018}: Wafer, Herring, and SwedishLeaf (details in \cite{liu2023timetransformer}). Previous studies argued oversampling approaches (via replication or simple modifications like jittering) can cause overfitting or create out-of-distribution synthesis \cite{yap2014over,yan2019otos}. Our model is expected to address these issues. Table \ref{augment} lists the classification results on the ECochG dataset using an off-the-shelf Multi-layer Perceptron (MLP) with (1) no augmentations, and augmenting the minority class via (2) repeating (3) jittering, (4) RTSGAN, (5) TimeVAE, and (6) our model (augmentations are only in training sets).
    \begin{table}[ht]
     \centering
     \caption{Imbalanced classification evaluation (ROC is receiver operating characteristic curve and PRC is precision-recall curve. Higher scores are better)}
     \scalebox{0.70}{%
     \begin{tabular}{cccccc}
         \hline
         Components & Accuracy & Recall & Precision & AUC\_ROC & AUC\_PRC \\ \hline\hline \xrowht{5pt}
         No Aug & 0.9751 & 0.7062 & 0.8706 & 0.9824 & 0.8521 \\
         Replication & 0.9719 & \textbf{0.9552} & 0.7072 & 0.9824 & 0.8636 \\
         Jittering & 0.9757 & 0.7074 & 0.8864 & 0.9866 & 0.8870 \\
         RTSGAN & 0.9702 & 0.6791 & 0.8235 & 0.9767 & 0.7352 \\
         TimeVAE & 0.9738 & 0.7388 & 0.8319 & 0.9807 & 0.7906 \\
         \textbf{Time-Trans} & \textbf{0.9802} & 0.8178 & \textbf{0.9163} & \textbf{0.9945} & \textbf{0.9345} \\ \hline
     \end{tabular}}
     \label{augment}
    \end{table}
    We do not list TimeGAN because training data here are insufficient to train this model. The results on the other three datasets can be found in \cite{liu2023timetransformer}.  One crucial problem of imbalanced classification is low recall rates, caused by an under-represented positive class in the training set. This can result in negative prediction dominance. Repeating the minority class can improve the recall at the cost of precision, which reflects the aforementioned overfitting problem. However, augmenting the training set with synthetic data from our model shows better results with respect to all metrics suggesting that it can aid in solving this kind of problem.

\begin{figure}[ht]
        \centering
        \includegraphics[scale=0.4]{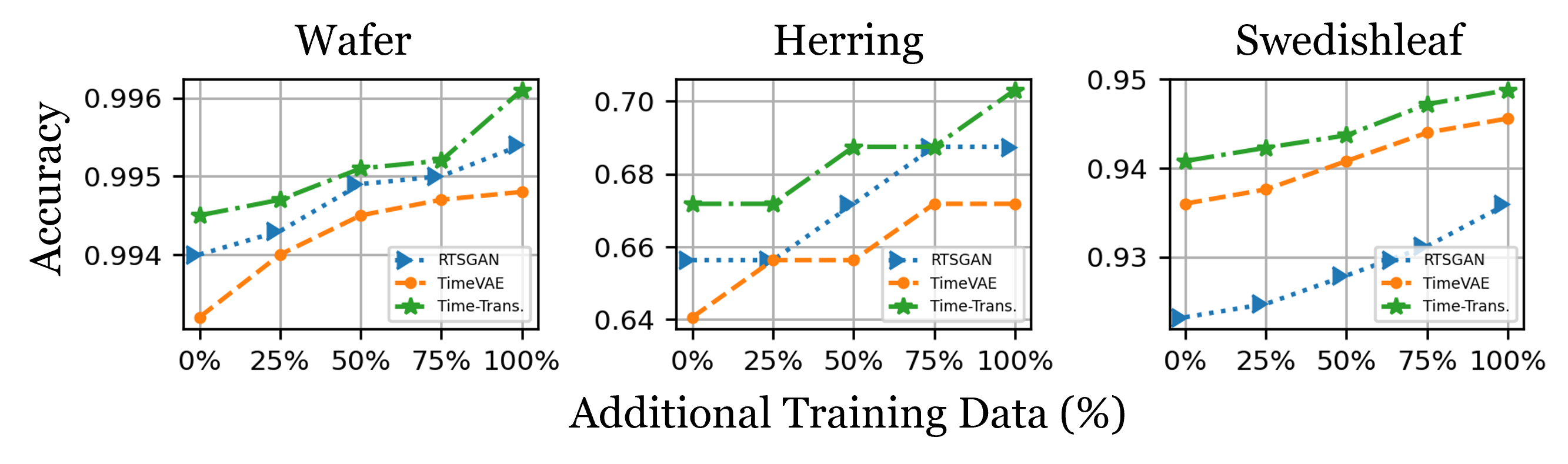}
        \caption{Classification results on augmented datasets with: Time-Trans(green), RTSGAN(blue), and TimeVAE(orange). Higher results are better.}
        \label{aug}
    \end{figure}

   Using the balanced datasets ($<$ 2000 time series), we investigate how 3 generative models can improve the deep learning training for small datasets. We extend these datasets with 25\%, 50\%, 75\%, and 100\% of their size and use them to train the same MLP. Figure \ref{aug} shows the change of accuracy and area under ROC and PRC curves.  The classifier performs better with more synthetic training data, and best with those from our model.
   %This highlights the potential of our model's utility.
   Detailed results are provided in \cite{liu2023timetransformer}.
   \vspace{-3mm}
   \subsubsection{Ablation Study \& Comparison with Sequential Design:}
    Using the ‘Music’ dataset, we investigate how each component of the model contributes to the final result. We (1) use only a De-convolutional block, (2) add only a TCN block after (1), (3) add only a Transformer block after (1), and (4) add our proposed Time-Trans after (1). Additionally, to test our claim that a parallel structure is better than a sequential combination, we run an extra experiment where we add a sequential combination of TCN and Transformer (Trans) after (1). Results are shown in Table \ref{ablation}(results of the other two local-global datasets are in \cite{liu2023timetransformer}). From the FID column we see a clear improvement when adding either a TCN or a Transformer. The sequential combination also improves the performance. Finally, our proposed parallel structure Time-Trans achieves an approximate 40\% improvement compared to the sequential one. The other two columns also reflect similar situations. These results indicate the strong ability of our proposed model to generate time series.
    \begin{table}[ht]
     \centering
     \caption{Ablation study. `Seq.' stands for Sequential. Lower results are better}
     \scalebox{0.72}{%
     \begin{tabular}{cccc}
        \hline \xrowht{5pt}
        Components & FID & Discriminative & Predictive \\ \hline\hline\xrowht{5pt}
        De-Convolution & 2.298±0.284 & 0.490±0.006 & 0.037±0.002 \\
        TCN & 0.672±0.097 & 0.471±0.013 & 0.026±0.007 \\
        Trans & 0.815±0.039 & 0.479±0.009 & 0.027±0.006 \\
        TCN+Trans (Seq.) & 0.627±0.031 & 0.455±0.021 & 0.022±0.008 \\
        Time-Trans & \textbf{0.395±0.013} & \textbf{0.160±0.054} & \textbf{0.017±0.009} \\ \hline
     \end{tabular}}
     \label{ablation}
    \end{table}
    \vspace{-2mm}

   % \subsubsection{Additional Experiments}
   %  We provide further investigations of our model with respect to: different choice of encoder, different training data size, local \& global feature extraction, longer time series data, and sophisticated evaluation models. Due to space limitations, we provide these results in supplementary file.

 \section{Limitations \& Border Impacts}
  Although we showed the strength of our model from multiple perspectives, we acknowledge there are some limitations and propose future work to mitigate these. First, as our model handles unsupervised generation, changes would need to be made for a conditional generation task, to produce user-defined synthetic time series data. Second, aiming at data augmentation, our model requires access to complete time series data. Expanding it to work with incomplete data and address time series imputation challenges would offer significant benefits. Apart from the positive impacts showed in \ref{apply}, our model may have possible negative social impact if used unethically (E.g. realistic fake time series data in news may be used to mislead people). It is important to follow ethical practices when using the model.

 \section{Conclusion}
  We introduced a novel time series generative model (Time-Transformer AAE), which contains an adversarial autoencoder (AAE) and a key component (Time-Transformer) in the decoder. Via a layer-wise parallelization and a bidirectional cross attention, Time-Transformer AAE exploits the learning ability of Temporal Convolutional Networks and Transformer in extracting local features and building global interaction respectively. We experimentally showed its effectiveness in generating time series data. Additionally, by training on different sizes of data, we showed the proposed model can achieve competitive performance to SOTA methods, even with less training data. Furthermore, we showed the model's utility in imbalanced classification and small dataset augmentation problems.\\

\noindent {\bf Acknowledgements:} This work is partially supported by NHMRC Development Grant Project 2000173 and The University of Melbourne’s Research Computing Services and the Petascale Campus Initiative. 

 \bibliographystyle{siam}
 \bibliography{ref}

\appendix
\onecolumn
 \section{ECochG dataset} \label{ecochgdes}
  Intra-operative electrocochleography (ECochG) is used to monitor the response of the cochlea to sound during a Cochlear Implant surgery. This surgery is a cost-effective solution for hearing impairment. However, a barrier to patients choosing to receive a cochlear implant is the high probability of losing their natural hearing due to trauma caused during the surgery. Previous studies show that the changes of some ECochG components can reflect such trauma \cite{campbell2016ecochg,dalbert2016assessment}. Specifically, \cite{campbell2016ecochg} demonstrate that a 30\% drop in amplitude of the `Cochlear Microphonic' (CM - one component of the ECochG signal) leads to poorer natural hearing preservation, and thus can be used to predict trauma. This has motivated researchers to develop machine learning models to automatically detect these drops and assist the surgeon in preventing trauma during the surgery \cite{wijewickrema2022automatic}.

 \section{Details of Imbalanced Datasets} \label{imbds}
  The original ECochG dataset contains data from 77 patients. When sampled and pre-processed as previously mentioned, this results in 13982 time series including two classes. However, this dataset is extremely imbalanced: only 874 instances are positive, which makes it hard to train a machine learning model. One possible solution is augmenting the minority class to make it balanced. As to the UCR datasets, the first two are imbalanced binary datasets while the last one contains 15 classes, we create an imbalanced situation by assigning class 1 as the positive class and all remaining classes as negative (labeled as 0). Here, we provide statistical details of the datasets used in \textbf{Model Application} of Section 4. Table \ref{imstats} shows this information including: number of training data (\#Train), number of testing data (\#Test), length of each time series (Len), positive rate of training set (Pos\_Rate, equals to \#Positive$/$\#Train), and number of training data for generative model training (\#G-Training).
  \begin{table}[ht]
      \centering
      \caption{Details of Imbalanced Datasets}
      \scalebox{0.85}{%
      \begin{tabular}{cccccc}
        \hline \xrowht{5pt}
          Dataset & \# Train & \# Test & Len & Pos\_Rate & \# G-Training \\ \hline\hline\xrowht{5pt}
          ECochG & 9787 & 4195 & 128 & 0.063 & 874 \\
          Wafer & 1000 & 6164 & 152 & 0.097 & 97 \\
          Herring & 64 & 64 & 512 & 0.391 & 25 \\
          SwedishLeaf & 500 & 625 & 128 & 0.058 & 29 \\ \hline
      \end{tabular}}
      \label{imstats}
  \end{table}

 \section{Experimental Setup} \label{expset}
   \subsection{Data Pre-processing}
    For the preliminary datasets, we follow the settings in previous work \cite{yoon2019timegan,desai2021timevae}. We use a sliding window of size 24 to sample the data resulting in time series of length 24, and re-scale all data to $[0,1]$ using the min-max normalization formula: $x' = \frac{x-min(x)}{max(x)-min(x)}$,
    where $x$ is the original data,and  $min(x), max(x)$ are minimum and maximum of the data respectively.

    We choose a longer length (128) for time series in local-global datasets, to preserve more local and global patterns, and to evaluate the model's ability to process longer data. We use min-max normalization to re-scale all the data to $[0,1]$ as mentioned previously.

   \subsection{Model Settings}
    We search the hyper-parameters of the autoencoder based on the average reconstruction performance on the preliminary datasets. We divide the datasets into training and validation sets, and search the hyper-parameters including layer number, filter number, kernel size, stride number, head number, head size, and dilation rate using the training set. The value range of each parameter is shown in Table \ref{hpval}.
    \begin{table}[ht]
        \centering
        \caption{Value range of Hyper-parameters}
        \scalebox{0.85}{%
        \begin{tabular}{cccccccc}
            \hline
            Hyper-parameters & layer number & filter number & kernel size & stride number & head number & head size & dilation rate \\ \hline\hline
            Value range & 2,3 & 64-512 & 4,6,8 & 1,2 & 2,3,4 & 64-256 & 1,2,4,16 \\ \hline
        \end{tabular}}
        \label{hpval}
    \end{table}
    As a result, the encoder is a 3-layer 1-dimensional CNN with 64, 128, and 256 filters and `relu' activation. All filters have kernels of size 4, and move 2 unit-step each time (stride of 2). After this convolutional process, the results are flattened and go through a fully connected layer which maps them into the latent code with pre-defined dimensions. Here we use 8-dimensional latent code for preliminary datasets (original: 24-step) and 16-dimensional vectors for local-global datasets (original: 128-step).

    In the decoder, the de-convolutional part has 2 transposed convolutional layers with 128 and 64 filters. All filters have kernels of size 4 and strides of 2 units. Then, the outputs are reshaped and mapped into the original data dimensions (length $L$ and channels $C$) via reshape and fully connected layers, which results in prototypes of the time series. The following Time-Transformer has two blocks, which represents a TCN that has two hidden dilated convolutional layers. The hidden layers have $C$ filters with kernel size of 4 and dilation rate of 1 and 4 respectively. The Transformer blocks combined with each dilated convolutional layer consist of a 3-head self-attention layer with head size of 64 and a feed-forward convolutional layer. The cross attention is also a 3-head attention module with a head size of 64. It takes inputs from different sides which makes it different from the self-attention module ahead. The discriminator, that accomplishes the adversarial process, is a 2-layer fully connected layer with 32 hidden units in both layers and `relu' activation.

    We follow the original AAE training procedure to train the model: We first update the encoder and decoder using the reconstruction loss (Mean Squared Error). Then, the encoder and discriminator are updated with respect to the adversarial loss consisting of a discriminative loss and a generative loss (both are cross-entropy). We use the Adam optimiser to update all the losses. The learning rate for reconstruction loss is 0.005 initially, and reduces to 0.0001 via a polynomial decay function (we directly implemented it using the tensorflow platform, the power of the polynomial is 0.5). Both the discriminative loss and generative loss have initial learning rates of 0.001, which also reduces to 0.0001 via a polynomial decay function (same as previous). The default training epoch is set to 1000. We provide the source code and a tutorial notebook here.

 % \section{Generation Examples}
 %  We show some generated samples (together with the original data) of local-global datasets in Figure \ref{sample}.
 %  \newpage
 %  \begin{figure}[h!]
 %      \centering
 %      \includegraphics[scale=0.42]{ori_music.png}
 %      \includegraphics[scale=0.42]{samp_music.png}\\
 %      \includegraphics[scale=0.42]{ori_sinecpx.png}
 %      \includegraphics[scale=0.42]{samp_sinecpx.png}\\
 %      \includegraphics[scale=0.42]{ori_ecochg.png}
 %      \includegraphics[scale=0.42]{samp_ecochg.png}
 %      \caption{Generation examples}
 %      \label{sample}
 %  \end{figure}

 \section{Additional Experimental Results}
   We provide more results from the experiments on visualization, training size, ablation study and model application as mentioned in the paper.

   \subsection{Visualization}\label{adexp}
\begin{figure}[ht]
      \centering
      \includegraphics[scale=0.35]{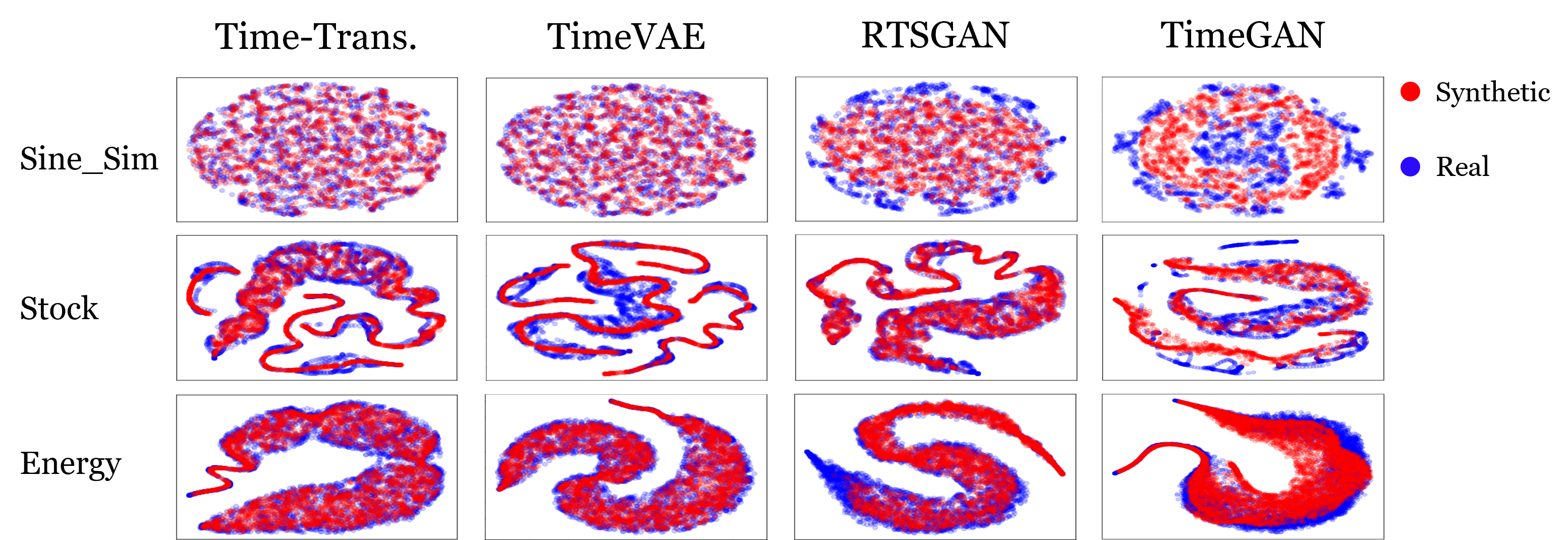}
      \caption{t-SNE visualizations on preliminary datasets}
      \label{tsnep}
    \end{figure}

    t-SNE plots of the preliminary datasets are shown in Figure \ref{tsnep}.

   \subsection{Longer time series data} \label{long_ts}
    Most existing related works \cite{yoon2019timegan,desai2021timevae,pei2021rtsgan} for synthetic time series generation have used 24-length time series as their experimental datasets. A recent work (PSA-GAN \cite{jeha2022psagan}) has claimed to generate long time series and the longest time series used in their work are 256-length. They apply a sampling method from LSTnet \cite{lai2018lstnet} to the electricity datasets, to produce these 256-length time series. Hence, we use this dataset, with the same sampling method as \cite{lai2018lstnet}, to evaluate our model. In addition, we further evaluate our model using another two real-world time series datasets from the UCR Archive \cite{UCRArchive2018}: Trace (cropped to 256-length) and ShapesAll (512-length). The results are shown in Table \ref{long1} below. We can see our model still performs well with these longer (256, 512-length) time series data, as it has the best performance against the baselines.
     (we only use the two best baseline models: RTSGAN and TimeVAE).
     \begin{table}[ht]
          \centering
          \caption{Long time series evaluation (lower is better for all metrics)}
          \scalebox{0.8}{%
          \begin{tabular}{ccccc}
              \hline
              Model & Dataset & FID & Disc. & Pred. \\ \hline\hline
              Oracle & \multirow{6}{*}{Electricity} & \multicolumn{2}{c}{N/A} & 0.010±0.004 \\
              TimeGAN &  & 9.501±1.654 & 0.488±0.016 & 0.105±0.014 \\
              TimeVAE &  & 1.655±0.621 & 0.415±0.037 & 0.034±0.005 \\
              RTSGAN &  & 0.896±0.147 & 0.397±0.007 & 0.021±0.007 \\
              Time-Trans. &  & \textbf{0.338±0.058} & \textbf{0.057±0.027} & \textbf{0.014±0.006} \\ \hline\hline
              Oracle & \multirow{6}{*}{Trace} & \multicolumn{2}{c}{N/A} & 0.062±0.001 \\
              TimgGAN &  & 17.011±1.859 & 0.499±0.003 & 0.229±0.014 \\
              TimeVAE &  & 7.024±0.228 & 0.482±0.015 & 0.116±0.004 \\
              RTSGAN &  & 7.959±0.217 & 0.497±0.006 & 0.175±0.002 \\
              Time-Trans. &  & \textbf{4.014±0.321} & \textbf{0.261±0.059} & \textbf{0.095±0.007} \\ \hline\hline
              Oracle & \multirow{6}{*}{ShapesAll} & \multicolumn{2}{c}{N/A} & 0.051±0.000 \\
              TimeGAN &  & 2.795±0.098 & 0.326±0.076 & 0.188±0.003 \\
              TimeVAE &  & 0.481±0.042 & 0.122±0.071 & 0.112±0.002 \\
              RTSGAN &  & 3.648±0.107 & 0.296±0.073 & 0.248±0.001 \\
              Time-Trans. &  & \textbf{0.445±0.029} & \textbf{0.034±0.045} & \textbf{0.094±0.005}\\ \hline
          \end{tabular}}
          \label{long1}
      \end{table}

    \subsection{Local \& Global Feature Learning details} \label{gl_feat}
     % First, we show some generated examples in Figure \ref{lg_gen}.
     % \begin{figure}[h!]
     %     \centering
     %     \begin{subfigure}[b]{1\textwidth}
     %      \centering
     %      \includegraphics[scale=0.44]{l_gen.png}
     %      \caption{Local feature series}
     %      \label{lts}
     %     \end{subfigure} \\
     %     \begin{subfigure}[b]{1\textwidth}
     %      \centering
     %      \includegraphics[scale=0.44]{25per.png}
     %      \caption{75\% Local feature}
     %      \label{25p}
     %     \end{subfigure}\\
     %     \begin{subfigure}[b]{1\textwidth}
     %      \centering
     %      \includegraphics[scale=0.44]{lg_gen.png}
     %      \caption{Mixture series}
     %      \label{lgts}
     %     \end{subfigure}\\
     %     \begin{subfigure}[b]{1\textwidth}
     %      \centering
     %      \includegraphics[scale=0.44]{75per.png}
     %      \caption{75\% Global feature}
     %      \label{75p}
     %     \end{subfigure}\\
     %     \begin{subfigure}[b]{1\textwidth}
     %      \centering
     %      \includegraphics[scale=0.44]{g_gen.png}
     %      \caption{Global feature series}
     %      \label{gts}
     %     \end{subfigure}
     %     \caption{Generation examples}
     %     \label{lg_gen}
     % \end{figure}
     % Comparing these generated examples, we can see: when the time series contain only global features or only local features, all the models can generate comparatively good time series (see Figure \ref{lts} and \ref{gts}). However, when it comes to the mixture of the two, baseline models fail to generate as effectively.

     Table \ref{eval_lg} lists the quantitative evaluations results of each model, which also shows the effectiveness of our model in extracting both global and local features simultaneously, as it gets the best scores with respect to all metrics for the mixture dataset. RTSGAN can capture the global trends but fails to extract local properties as it performs well when data contains more global features. On the other hand, TimeVAE is good at local features learning according to the results. TimeGAN and RCGAN seem to be able to learn a single feature to some extent, but they fail to capture the mixed features.
     \begin{table*}[ht]
      \centering
      \caption{Evaluation results for global \& local features learning (lower scores are better)}
      \scalebox{0.85}{%
      \begin{tabular}{ccccccc}
        \hline
        Model & Benchmark & Local & 25\%G & 50\%G & 75\%G & Global \\ \hline\hline
        TimeGAN & \multirow{4}{*}{FID} & 4.736±0.224 & 5.274±0.561 & 5.195±0.316 & 6.139±0.738 & 1.597±0.294 \\
        RTSGAN &  & 0.299±0.068 & 0.823±0.042 & 1.788±0.091 & 0.462±0.018 & \textbf{0.163±0.055} \\
        TimeVAE &  & \textbf{0.271±0.074} & 0.597±0.054 & 1.081±0.068 & 0.934±0.076 & 0.241±0.037 \\
        Time-Trans. &  & 0.317±0.071 & \textbf{0.463±0.038} & \textbf{0.474±0.014} & \textbf{0.441±0.052} & 0.229±0.019 \\ \hline\hline
        TimeGAN & \multirow{4}{*}{Disc.} & 0.178±0.062 & 0.476±0.028 & 0.485±0.011 & 0.495±0.006 & 0.048±0.007 \\
        RTSGAN &  & 0.022±0.019 & 0.067±0.037 & 0.294±0.078 & \textbf{0.062±0.049} & \textbf{0.012±0.004} \\
        TimeVAE &  & \textbf{0.015±0.026} & \textbf{0.051±0.040} & 0.177±0.073 & 0.183±0.045 & 0.018±0.009 \\
        Time-Trans. &  & 0.029±0.027 & 0.055±0.035 & \textbf{0.061±0.060} & 0.063±0.051 & 0.017±0.015 \\ \hline\hline
        Oracle & \multirow{5}{*}{Pred.} & 0.009±0.002 & 0.014±0.003 & 0.024±0.003 & 0.021±0.007 & 0.014±0.008 \\
        TimeGAN &  & 0.077±0.018 & 0.098±0.027 & 0.103±0.043 & 0.106±0.011 & 0.052±0.013 \\
        RTSGAN &  & 0.014±0.006 & 0.046±0.013 & 0.073±0.015 & 0.034±0.018 & \textbf{0.027±0.012} \\
        TimeVAE &  & \textbf{0.013±0.006} & 0.031±0.007 & 0.052±0.011 & 0.058±0.009 & 0.031±0.005 \\
        Time-Trans. &  & \textbf{0.013±0.005} & \textbf{0.026±0.007} & \textbf{0.031±0.007} & \textbf{0.033±0.011} & 0.030±0.009 \\ \hline
      \end{tabular}}
      \label{eval_lg}
     \end{table*}

  \subsection{Imbalanced Classification} \label{imc}
   Table \ref{ucraugment} lists imbalanced classification results on UCR datasets: Wafer, Herring, and SwedishLeaf.
   \begin{table*}[ht]
    \centering
    \caption{Imbalanced Classification Evaluation for UCR datasets (higher results are better)}
    \scalebox{0.79}{%
    \begin{tabular}{ccccccc}
        \hline
        Datasets & Components & Accuracy & Recall & Precision & AUC\_ROC & AUC\_PRC \\ \hline\hline \xrowht{5pt}
        \multirow{6}{*}{Wafer} & No Aug & 0.9919 & 0.9639 & 0.9610 & 0.9981 & 0.9826 \\
        & Replication & 0.9924 & 0.9835 & 0.9478 & 0.9979 & 0.9855 \\
        & Jittering & 0.9933 & 0.9774 & 0.9643 & 0.9985 & 0.9892 \\
        & RTSGAN & 0.9940 & 0.9759 & \textbf{0.9686} & 0.9973 & 0.9847 \\
        & TimeVAE & 0.9932 & 0.9789 & 0.9587 & 0.9978 & 0.9868 \\
        & \textbf{Time-Trans.} & \textbf{0.9945} & \textbf{0.9849} & 0.9618 & \textbf{0.9992} & \textbf{0.9928} \\ \hline\hline \xrowht{5pt}
        \multirow{6}{*}{Herring} & No Aug & 0.5469 & 0.3461 & 0.5625 & 0.6549 & 0.5762 \\
        & Replication & 0.6563 & \textbf{0.4615} & 0.6000 & 0.7287 & 0.6429 \\
        & Jittering & 0.6250 & 0.3846 & 0.5556 & 0.6781 & 0.5455 \\
        & RTSGAN & 0.6563 & 0.4231 & 0.6111 & 0.7146 & 0.6307 \\
        & TimeVAE & 0.6406 & 0.3846 & 0.5882 & 0.6817 & 0.5891 \\
        & \textbf{Time-Trans.} & \textbf{0.6875} & \textbf{0.4615} & \textbf{0.6667} & \textbf{0.7712} & \textbf{0.6455} \\ \hline\hline \xrowht{5pt}
        \multirow{6}{*}{SwedishLeaf} & No Aug & 0.9264 & 0.0000 & 0.0000 & 0.9383 & 0.4296 \\
        & Replication & 0.8848 & 0.8043 & 0.3700 & 0.9487 & 0.5337 \\
        & Jittering & 0.9279 & 0.3043 & 0.5185 & 0.9426 & 0.5230 \\
        & RTSGAN & 0.9232 & 0.1956 & 0.4500 & 0.9418 & 0.4656 \\
        & TimeVAE & 0.9360 & 0.6304 & 0.5577 & 0.9439 & 0.5433 \\
        & \textbf{Time-Trans.} & \textbf{0.9408} & \textbf{0.8478} & \textbf{0.5652} & \textbf{0.9536} & \textbf{0.6341} \\ \hline
     \end{tabular}}
     \label{ucraugment}
    \end{table*}

  \subsection{Small Dataset Augmentation} \label{sml_aug}
   We provide the detailed results on small datasets augmentation in Table \ref{smlaug}.
   \begin{table*}[ht]
     \centering
     \caption{Data Augmentation Evaluation (higher results are better)}
     \scalebox{0.80}{%
     \begin{tabular}{ccccccccccc}
      \hline
      \multicolumn{2}{c}{Dataset} & \multicolumn{3}{c}{Wafer} & \multicolumn{3}{c}{Herring} & \multicolumn{3}{c}{Swedisleaf} \\ \cline{3-11}
      \multicolumn{2}{c}{\multirow{2}{*}{Model}} & \multirow{2}{*}{RTSGAN} & \multirow{2}{*}{TimeVAE} & \multirow{2}{*}{\begin{tabular}[c]{@{}c@{}}Time-\\ Trans.\end{tabular}} & \multirow{2}{*}{RTSGAN} & \multirow{2}{*}{TimeVAE} & \multirow{2}{*}{\begin{tabular}[c]{@{}c@{}}Time-\\ Trans.\end{tabular}} & \multirow{2}{*}{RTSGAN} & \multirow{2}{*}{TimeVAE} & \multirow{2}{*}{\begin{tabular}[c]{@{}c@{}}Time-\\ Trans.\end{tabular}} \\
      \multicolumn{2}{c}{} &  &  &  &  &  &  &  &  &  \\ \hline
      Accuracy & +0\% & 0.9940 & 0.9932 & \textbf{0.9945} & 0.6563 & 0.6406 & \textbf{0.6718} & 0.9232 & 0.9360 & \textbf{0.9408} \\
       & +25\% & 0.9943 & 0.9940 & \textbf{0.9947} & 0.6563 & 0.6563 & \textbf{0.6718} & 0.9247 & 0.9376 & \textbf{0.9423} \\
       & +50\% & 0.9949 & 0.9945 & \textbf{0.9951} & 0.6718 & 0.6563 & \textbf{0.6875} & 0.9279 & 0.9408 & \textbf{0.9437} \\
       & +75\% & 0.9950 & 0.9947 & \textbf{0.9952} & 0.6875 & 0.6718 & \textbf{0.6875} & 0.9312 & 0.9440 & \textbf{0.9472} \\
       & +100\% & 0.9954 & 0.9948 & \textbf{0.9961} & 0.6875 & 0.6718 & \textbf{0.7031} & 0.9360 & 0.9456 & \textbf{0.9488} \\ \hline
      \multirow{5}{*}{AUC\_ROC} & +0\% & 0.9973 & 0.9978 & \textbf{0.9987} & 0.7146 & 0.6817 & \textbf{0.7712} & 0.9418 & 0.9439 & \textbf{0.9536} \\
       & +25\% & 0.9980 & 0.9979 & \textbf{0.9989} & 0.7256 & 0.7086 & \textbf{0.7758} & 0.9415 & 0.9526 & \textbf{0.9528} \\
       & +50\% & 0.9985 & 0.9982 & \textbf{0.9990} & 0.7316 & 0.7138 & \textbf{0.7828} & 0.9464 & 0.9532 & \textbf{0.9578} \\
       & +75\% & 0.9988 & 0.9985 & \textbf{0.9991} & 0.7398 & 0.7378 & \textbf{0.7891} & 0.9553 & 0.9582 & \textbf{0.9609} \\
       & +100\% & 0.9990 & 0.9986 & \textbf{0.9993} & 0.7537 & 0.7419 & \textbf{0.7936} & 0.9569 & 0.9603 & \textbf{0.9628} \\ \hline
      \multirow{5}{*}{AUC\_PRC} & +0\% & 0.9847 & 0.9868 & \textbf{0.9928} & 0.6307 & 0.5891 & \textbf{0.6455} & 0.4656 & 0.5933 & \textbf{0.6341} \\
       & +25\% & 0.9856 & 0.9869 & \textbf{0.9929} & 0.6367 & 0.5936 & \textbf{0.6479} & 0.4827 & \textbf{0.6536} & 0.6435 \\
       & +50\% & 0.9866 & 0.9869 & \textbf{0.9932} & 0.6420 & 0.5982 & \textbf{0.6505} & 0.5324 & 0.6548 & \textbf{0.6642} \\
       & +75\% & 0.9876 & 0.9870 & \textbf{0.9935} & 0.6497 & 0.6001 & \textbf{0.6628} & 0.5922 & 0.6934 & \textbf{0.6993} \\
       & +100\% & 0.9896 & 0.9873 & \textbf{0.9937} & 0.6603 & 0.6129 & \textbf{0.6718} & 0.6126 & 0.7330 & \textbf{0.7460} \\ \hline
     \end{tabular}}
     \label{smlaug}
   \end{table*}

    \subsection{Choice of Encoder} \label{enc_chs}
     Table \ref{enc} lists different performance outcomes from our model with two different encoders, namely, CNN and Time-Transformer. The first is the simple design which we used in our proposed model, and the second is the one with a Time-Transformer module inserted in the encoder. We test their performance on the Sine\_Sim dataset to briefly investigate how much improvement a complex encoder can bring to the model. As shown in Table \ref{enc}, using a Time-Transformer encoder improves the results a little, but it requires much more time to train. This indicates that a simple encoder can generate relatively good synthetic data with high efficiency.
     \begin{table}[h!]
        \centering
        \caption{Comparison between different encoders (lower results are better)}
        \begin{tabular}{ccccc}
          \hline
          Encoder & Training Time (s) & FID & Discriminative & Predictive \\ \hline \hline
          CNN & \textbf{530.24} & 0.283±0.023 & 0.131±0.021 & 0.051±0.015 \\
          Time-Transformer & 1651.33 & \textbf{0.256±0.014} & \textbf{0.120±0.011} & \textbf{0.036±0.010} \\ \hline
        \end{tabular}
        \label{enc}
      \end{table}

    \subsection{Additional ablation study} \label{ad_ablation}
   Table \ref{ablation_ad} shows the results of ablation study with respect to `Sine\_Cpx' and `ECochG' datasets. Both show performance improvements when we add each part.
      \begin{table}[ht]
          \centering
          \caption{Extra Ablation Study}
          \scalebox{0.9}{%
          \begin{tabular}{ccccc}
              \hline
              Dataset & Components & FID & Discriminator & Predictor \\ \hline\hline
              \multirow{5}{*}{Sine\_Cpx} & De-Conv & 6.976±0.157 & 0.490±0.021 & 1.832±0.526 \\
              & TCN & 2.570±0.067 & 0.417±0.270 & 0.529±0.057 \\
              & Transformer & 2.472±0.079 & 0.397±0.253 & 0.484±0.025 \\
              & TCN+Trans (Sequential) & 1.721±0.055 & 0.193±0.181 & 0.037±0.013 \\
              & Time-Transformer & 1.502±0.062 & 0.168±0.041 & 0.032±0.006 \\ \hline\hline
              \multirow{5}{*}{ECochG} & De-Conv & 0.781±0.099 & 0.461±0.048 & 0.092±0.011 \\
              & TCN & 0.502±0.041 & 0.288±0.192 & 0.027±0.013 \\
              & Transformer & 0.513±0.045 & 0.256±0.164 & 0.027±0.007 \\
              & TCN+Trans (Sequential) & 0.402±0.031 & 0.162±0.059 & 0.017±0.005 \\
              & Time-Transformer & 0.348±0.024 & 0.104±0.012 & 0.013±0.006 \\ \hline
          \end{tabular}}
          \label{ablation_ad}
      \end{table}

  \subsection{Computation Details}
   We ran all the models on the three preliminary datasets and recorded their running time. The results are provided in Table \ref{timecomp} (all models were run on High Performance Computer with Intel(R) Xeon(R) Gold 6132 CPU and v100 GPU, values are formatted in seconds):
   \begin{table}[ht]
     \centering
     \caption{Running Time Comparison}
     \begin{tabular}{cccc}
      \hline
      Models & Sine\_Sim & Stock & Energy \\ \hline\hline
      TimeGAN & 2589.16 & 2870.19 & 4776.35 \\
      TimeVAE & 183.26 & 286.52 & 1301.81 \\
      RTSGAN & 810.49 & 1298.74 & 2433.31 \\
      \textbf{Time-Trans.} & 176.69 & 299.36 & 1313.47 \\ \hline
     \end{tabular}
     \label{timecomp}
   \end{table}

   For the memory comparison, we list the trainable parameters of each model in the Table \ref{memcomp} (the parameters were calculated with input size of (24, 1)).
    \begin{table}[h!]
        \centering
        \caption{Trainable Parameters Comparison}
        \begin{tabular}{ccccc}
            \hline
            Models & TimeGAN & TimeVAE & RTSGAN & \textbf{Time-Trans.}\\ \hline\hline
            Parameters & 70712 & 266893 & 64530 & 217942 \\ \hline
        \end{tabular}
        \label{memcomp}
    \end{table}

   From the two tables, we can see our model and TimeVAE are much faster than other models though they have more parameters. It is because most of the parameters of these two models are based on CNNs where parameters can be processed in parallel, while other models are mostly using RNNs which are much slower than CNNs because they can only process sequentially. (please see \url{https://github.com/baidu-research/DeepBench} for a detailed comparison between these two architectures). Thus, even though we designed a complex model containing more parameters than RNN-based models, it can still be faster.

 \subsection{Justification of Baseline Models} \label{justify}
  As mentioned in the main part of the paper, we calculate the Predictive Score in a different way to the related works. We used the one-step-ahead forecast. Assume we have a time series $x=[t_1, \cdots, t_n ]$, where $t_i = [t_i^{(1)}, \cdots, t_i^{(C)}] ^T$. The inputs are the first $n-1$ time steps of all dimensions ($x_{train} = [t_1, \cdots, t_{n-1} ] $), and the model learns to predict the last time step ($y_{train} = [ t_n ]$). The loss is the mean absolute error between the predictions and the ground truth ($| y_{train} - y_{pred} |$). However, most related works directly use the method from TimeGAN. There, the inputs to the predictor are the first $n-1$ steps of the first $C-1$ dimensions ($x_{train} = [t_1', \cdots, t_{n-1}' ]$, where $t_i' = [ t_i^{(1)}, \cdots, t_i^{(C-1)} ]^T$). The model learns to predict the last dimension with one more step ($y_{train} = [ t_2^{(C)}, \cdots, t_n^{(C)} ]$). The loss is the mean absolute error with respect to the new outputs. Futhermore, TimeGAN used a sigmoid function as the last activation (while we don't use any activation in the last layer).

  These methodological differences led to the different results of the Predictive Score between our work and that of related works. Now, we have also used their method to get a new set of Predictive Scores. The results are shown in Table \ref{justpred} below.
  \begin{table}[h!]
    \centering
    \caption{Predictive Score using the same method in TimeGAN (lower scores are better)}
    \scalebox{0.85}{%
    \begin{tabular}{ccccccc}
      \hline
      Model & Sine\_sim & Stock & Energy & Music & Sine\_Cpx & ECochG \\ \hline\hline
      Oracle & 0.215±0.001 & 0.037±0.000 & 0.249±0.002 & 0.080±0.001 & 0.168±0.003 & 0.008±0.000 \\
      TimeGAN & 0.257±0.004 & 0.040±0.001 & 0.278±0.005 & 0.106±0.003 & 0.234±0.005 & 0.013±0.002 \\
      TimeVAE & \textbf{0.218±0.001} & 0.040±0.000 & 0.254±0.001 & 0.102±0.002 & 0.189±0.001 & 0.011±0.003 \\
      RTSGAN & 0.242±0.002 & \textbf{0.038±0.000} & \textbf{0.253±0.005} & 0.097±0.001 & 0.227±0.006 & 0.014±0.002 \\
      Time-Trans. & 0.223±0.002 & 0.039±0.001 & 0.261±0.001 & \textbf{0.086±0.001} & \textbf{0.187±0.001} & \textbf{0.009±0.000} \\
      \hline
    \end{tabular}}
    \label{justpred}
  \end{table}
  These results are similar to the related works on the corresponding datasets, taking into account acceptable variances caused by random seeds and hardware environments. This provides additional evidence that our baseline models were implemented properly. On the other hand, we can see although the values have changed, our claim in the paper still stands: our model performs the best for the three local-global datasets (Music, Sine\_Cpx, ECochG), which indicates its effectiveness in extracting global and local features simultaneously.

\end{document}